\documentclass{article}
\usepackage{spconf,amsmath,graphicx}
\usepackage{booktabs} 
\usepackage{color} 
\usepackage[table]{xcolor}
\usepackage{graphicx}
\usepackage{url}
\usepackage{amssymb}

\usepackage{enumitem}
\usepackage{comment}
\setlist{nosep, leftmargin=14pt}

\usepackage{mwe} 

\usepackage{pifont}
\newcommand{\cmark}{\ding{51}}%
\newcommand{\xmark}{\ding{55}}%

\definecolor{pink}{RGB}{239,189,216}
\definecolor{burgundy}{RGB}{201,123,154}

\title{Exploiting Spatial Structure for Transductive Few-Shot Classification of Whole-Slide Images}
\name{
\parbox{\linewidth}{\centering
Tiffanie Godelaine$^{\star}$ \hspace{3mm} Manon Dausort$^{\star}$ \hspace{3mm} \hspace{3mm} Karim El Khoury \hspace{3mm} Beno\^it Gérin\\
Beno\^it Macq \hspace{4mm} Christophe De Vleeschouwer \thanks{$^\star$The authors have contributed equally to this work.}}}

\address{ICTEAM, Université Catholique de Louvain, Belgium}

\begin{document}

\maketitle

\begin{abstract}
Automating the analysis of whole-slide images (WSIs), a key step in cancer diagnosis, has high clinical value, as it can reduce pathologist's workload while improving diagnosis accuracy. Recently, vision-language models have shown promising performance for patch-level classification without requiring any annotation, yet these zero-shot (ZS) predictions remain noisy on fine-grained tasks and must be further refined. 
A promising direction is to refine all predictions jointly, i.e., a transductive approach. However, most existing methods are not tailored to WSIs. We thus propose SlideTIM, an adaptation to WSIs of the recent transductive approach LC-TIM, which introduces a combined spatial--latent regularizer together with a prior on the patch class distribution. The former enforces spatially and semantically close patches to receive the same predictions, while the prior calibrates the predicted class proportions. Together, they address the complex spatial organization and the strong class imbalance of WSIs.
Evaluated on four histology datasets, SlideTIM consistently outperforms all TIM variants, improving the macro-F1 by +8.1pp over the best competing baseline at 1 shot. Compared to the ZS, it raises the macro-F1 by +19.4pp at 1 shot. The code will be made available after submission.
\end{abstract}


\section{Introduction}
\label{sec:intro}

Histopathological whole-slide images (WSIs) enable the precise visualization of lesions, which is crucial to characterize them as benign or malignant and, when malignant, to grade and stage the cancer.
However, owing to the very large size of these images, their analysis is time-consuming for pathologists~\cite{Borowsky2020}.
Deep learning can help reduce this workload while improving diagnostic accuracy by classifying the patches that compose the slide~\cite{Hekler2019}.

Recently, this task has increasingly relied on histology vision-language models (VLMs)~\cite{Huang2023, Ikezogwo2023, Lu2024conch}. By exploiting text supervision, these models yield strong patch-level zero-shot (ZS) predictions, requiring no manual annotation. Nevertheless, such ZS predictions remain noisy on fine-grained histology tasks and must be refined with a few annotated patches~\cite{AlMajzoub2026}.

The methods that refine these patch-level predictions from few annotations can be divided into two categories.
First, \textit{inductive} few-shot methods refine the prediction of each patch independently~\cite{Shakeri2024, Meseguer2025}.
Second, \textit{transductive} few-shot methods refine the predictions of all patches jointly by exploiting the relations between them~\cite{Sadraoui2024paddle, Zanella2025histotransclip, Godelaine2026histocrf, Godelaine2026slidecrf}.
The latter direction has attracted growing interest in histology, since the patches of a given slide are strongly related, making the exploitation of these relations particularly beneficial. 

Histo-TransCLIP~\cite{Zanella2025histotransclip}, an adaptation of TransCLIP~\cite{Zanella2024transclip} to histology, refines the predictions in the latent space by modelling the data distribution as a Gaussian mixture. HistoCRF~\cite{Godelaine2026histocrf} adapts conditional random fields to histology, leveraging few annotations and inter-patch similarities to refine the VLM predictions. However, these methods are evaluated on sets of independent patches. More recently, SlideCRF~\cite{Godelaine2026slidecrf}, built upon HistoCRF, incorporates spatial relations to handle WSIs and demonstrates the importance of this information. 

In parallel, LC-TIM~\cite{Khoury2026lctim}, a variant of the transductive method TIM++~\cite{Li2026tim++}, has been introduced. 
TIM++ adds supervision from VLMs to TIM~\cite{Dolz2020tim}. 
This latter formulates classification as jointly fitting a classifier to a few labeled examples, while maximizing the mutual information between the features of the unlabeld images and their predicted labels.
LC-TIM extends the objective function of TIM++ with a latent-space local-consistency regularizer, enforcing that samples close to one another in the embedding space receive the same prediction.
\vspace{1mm}

\begin{figure}[h!]
    \centering
    \includegraphics[width=\linewidth]{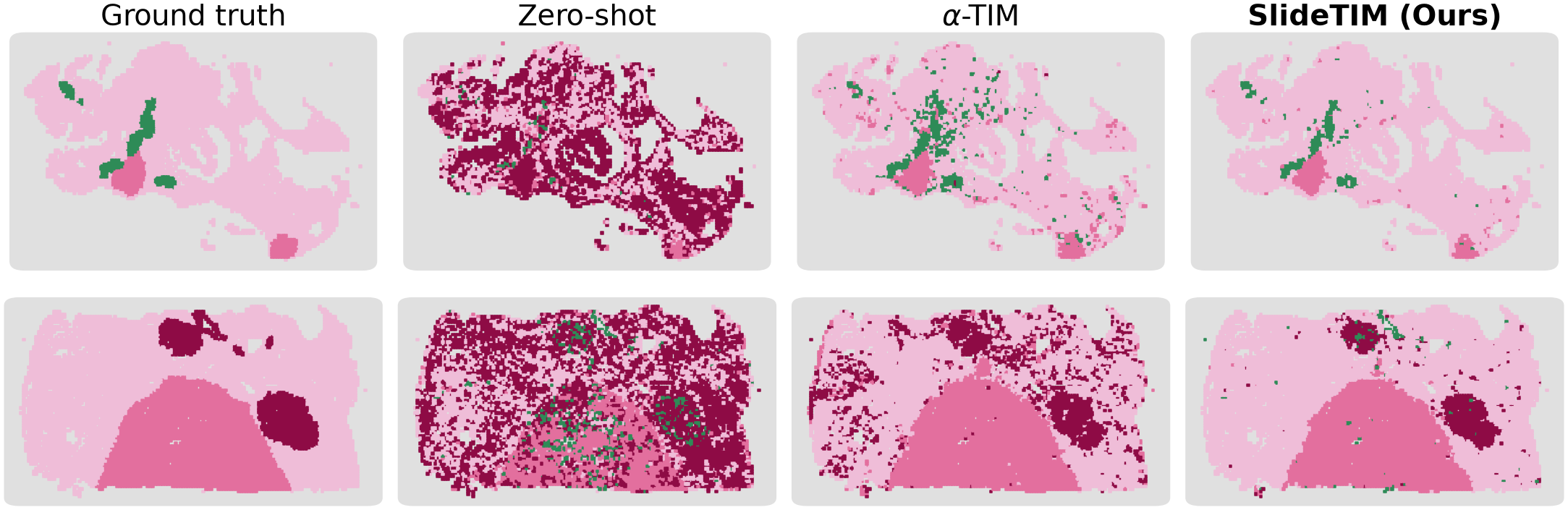}
    \caption{Examples of the refinement result of the ZS prediction using our method on two slides, compared to the best baseline. SlideTIM allows for finer fragmentation of structures within a slide.}
    \label{fig:visual_results}
\end{figure}

In this work, we adapt LC-TIM to the histology WSI setting, which is characterized by a complex spatial organization and a high class imbalance. This leads to two \textbf{contributions}:
\begin{itemize}
    \item We propose \textbf{SlideTIM}, which extends LC-TIM with a combined spatial--latent regularizer and a prior on the patch class distribution, making the method robust to the regions that are widespread within a WSI and to the strong class imbalance.  As shown in Figure \ref{fig:visual_results}, the refinement recovers the regions in the slide, while being robust to class imbalance.
    \item We show that it outperforms existing TIM variants by \textbf{+8.3}pp in macro-F1 at one shot per class present in the slide. The same tendency is observed when the number of shot increases.Compared to ZS, it improves the macro-F1 by \textbf{+19.4}pp and \textbf{+34.6}pp at 1 and 16 shots. 
\end{itemize}

\section{Method}
\label{sec:method}

\subsection{Problem setting}
We address few-shot classification of histopathology patches from a WSI in a transductive setting. For a given WSI, we dispose of a small labeled set $S=\{(\mathbf{f}_i,y_i)\}$ (the few-shot patches, with a label $y_i\in\{1,\dots,K\}$) and a large unlabeled set $Q$ made of all remaining patches of the slide.
Each patch $i$ is encoded by a frozen pathology vision encoder into an embedding $\mathbf{f}_i \in \mathbb{R}^{d}$. 
We use both sets to optimize a soft linear classifier $\mathbf{W}=[\mathbf{w}_1,\dots,\mathbf{w}_K]\in\mathbb{R}^{d\times K}$ that assigns to each patch $i$ the prediction
\begin{equation}
    p_{ik} = \frac{\exp\!\big(-\tfrac{\tau}{2}\,\| \mathbf{f}_i - \mathbf{w}_k \|^2\big)}
                  {\sum_{j=1}^{K}\exp\!\big(-\tfrac{\tau}{2}\,\| \mathbf{f}_i - \mathbf{w}_j \|^2\big)},
\label{eq:pred}
\end{equation}
with $\tau$ a temperature parameter and $K$ the number of classes. 

\subsection{Baseline: LC-TIM objective function}
The starting point is a transductive information-maximization method \cite{Khoury2026lctim} that combines supervision on $S$ with a mutual information term on $Q$, a text-based regularizer, and a latent-space regularizer:
\begin{multline}
    \min_{\textbf{W}}
    \underbrace{\lambda\,\mathrm{CE}(\textbf{W}, S)}_{\text{Support supervision}}
  -  \underbrace{(\hat{H}(Y_Q) - \alpha \hat{H}(Y_Q \mid X_Q))}_{\text{Mutual information}} \\
  +  \underbrace{\mathrm{KL}_\beta (p \,\|\, \hat y)}_{\text{Text regularization}}
  + \underbrace{\delta\,\mathcal{R}(p)}_{\text{Latent-space regularizer}},
\label{eq:objective_lctim}
\end{multline}
where $Y_Q$ and $X_Q$ denote the label and input distributions in the query set, respectively. $\alpha$, $\beta$, $\lambda$, $\delta$ are weighting factors. \vspace{1mm}
\begin{itemize}
    \item \textbf{Support supervision:} The cross-entropy $\mathrm{CE}(\textbf{W}, S)$ term fits the classifier to the few labeled patches. \vspace{1mm}
    \item \textbf{Mutual information:} Maximizing the marginal entropy $\hat H(Y_Q)$ while minimizing the conditional entropy $\hat H(Y_Q\mid X_Q)$ yields predictions that are individually confident and marginally balanced over classes. The marginal term implicitly pushes the class distribution $\overline{p} = \frac{1}{|Q|}\sum_{i \in Q}p_i$ towards the uniform $\mathbf{u}$. \vspace{1mm}
    \item \textbf{Text regularization:} Minimizing the Kullback-Leibler (KL) divergence $\mathrm{KL}_\beta(p \,\|\, \hat y) = \frac{1}{|Q|} \sum_{i \in Q} \mathrm{KL}_\beta(p_i \,\|\, \hat y_i)$ keeps the prediction $p_{i}$ close to the ZS prediction $\hat{y}_{i}$, obtained with a VLM: 
    \begin{equation}
        \hat{y}_{ik} = \frac{\exp (\tau \mathbf{f}_i^\top \mathbf{t}_k)}{\sum_{j=1}^K \exp (\tau \mathbf{f}_i^\top \mathbf{t}_j)}.
    \end{equation}
    where $\mathbf{f}_i$ is obtained by encoding each patch $i$ with the visual encoder of the VLM; and $\mathbf{t}_k$ is obtained by averaging a set of prompts (e.g., \textit{``a histology image of \{class\}''}) for each class $k$~\cite{Zanella2025histotransclip}. 
    Here $\mathrm{KL}_\beta$ denotes that only the cross-entropy component of the divergence is weighted by $\beta$; its entropy component $-H(p_i)$ is already accounted for by the conditional-entropy term.
    \vspace{1mm}
    \item \textbf{Latent-space regularizer: } The latent-space regularizer
    \begin{equation}
        \mathcal{R}(p) = \frac{1}{|Q|} \sum_{i \in Q} \mathrm{KL}\!\big(p_i \,\big\|\, \bar p_i \big),
        \qquad \bar p_i = \frac{1}{|\mathcal{N}_i|} \sum_{j \in \mathcal{N}_i} p_j,
    \end{equation}
    encourages each patch $i$ to agree with its neighbors $\mathcal{N}_i= \operatorname{top-}\!\kappa(A_{i\cdot})$, with $\kappa$ the number of neighbors. $\mathcal{N}_i$ gathers the most similar patches according to the latent-space similarity
    $a_{ij} = \mathbf{f}_i^\top\mathbf{f}_j$,
    where $\mathbf{f}$ is the normalized embedding obtained with a histology visual encoder.
\end{itemize}

\subsection{SlideTIM objective function}
We extend the baseline in two ways: a \textbf{prior-anchored marginal entropy} and a \textbf{spatial--latent regularizer}: \vspace{1mm}
\begin{itemize}
\item \textbf{Prior-anchored marginal entropy:}
    The key observation is that maximizing the marginal entropy is nothing but minimizing a KL divergence to the uniform distribution $\mathbf{u}$ \cite{Veilleux2021alphatim}: 
    \begin{equation}
        \hat H(Y_Q) = \log K - \mathrm{KL}\!\big(\bar p \,\|\, \mathbf{u}\big).
    \label{eq:entropy_as_kl}
    \end{equation}
    We therefore simply replace the reference distribution $\mathbf{u}$ by an estimated per-slide prior $\hat\pi$:
    \begin{equation}
        H_{\hat\pi}(Y_Q) \;:=\; -\,\mathrm{KL}\!\big(\bar p \,\|\, \hat\pi\big)
        \;=\; \hat H(Y_Q) + \langle \bar p,\, \log \hat\pi \rangle ,
    \label{eq:prior_entropy}
    \end{equation}
    This pulls the predicted class marginal $\bar p$ towards the prior $\hat\pi$.
    We estimate $\hat\pi$ from the labeled set: we compute one centroid per class as the mean embedding, provided by a visual encoder, of the labeled patches of that class, assign each unlabeled patch to its nearest centroid, and take $\hat\pi$ as the normalized histogram of these assignments.
    \vspace{1mm}
    \item \textbf{Spatial--latent regularizer:} Instead of a neighborhood defined in the latent space only, we merge the embedding and the spatial cues into a single affinity, so that patches close in both spaces are encouraged to share the same prediction. Let $\mathbf{x}_i$ be the spatial coordinates of patch $i$; the affinity is the product of a latent-space term $a^u$ and a spatial term $a^s$:
    \begin{equation}
        a_{ij} = a^{u}_{ij}\cdot a^{s}_{ij},
        \quad
        \begin{cases}
            a^u_{ij}=\tilde{\mathbf{f}}_i^\top\tilde{\mathbf{f}}_j,\\[2pt]
            a^s_{ij}=e^{-\|\mathbf{x}_i-\mathbf{x}_j\|^2/2\sigma_s^2},
        \end{cases}
    \label{eq:affinity}
    \end{equation}
    where $a^u$ is the cosine similarity in the embedding space and $a^s$ a spatial Gaussian kernel of bandwidth $\sigma_s$. Each affinity is min--max normalized row-wise, so that both cues have a comparable scale before being multiplied. 
    \vspace{1mm}
\end{itemize}
Finally, substituting $\hat H(Y_Q)\!\to\! H_{\hat\pi}(Y_Q)$ in Eq.~\eqref{eq:objective_lctim} and adding the spatial component from Eq.~\ref{eq:affinity} to the latent-space regularizer yield the SlideTIM objective:
\begin{multline}
    \min_{\textbf{W}}
    \lambda\,\mathrm{CE}(\textbf{W}, S)
  - \underbrace{\big(\,H_{\hat\pi}(Y_Q) - \alpha \hat{H}(Y_Q \mid X_Q)\big)}_{\text{Prior-anchored mutual information}} \\
  + \mathrm{KL}_\beta(p \,\|\, \hat y)
  + \underbrace{\delta\,\mathcal{R}(p)}_{\text{Spatial--latent regularizer}}.
    \label{eq:objective_ours}
\end{multline}
Setting $\hat\pi=\mathbf{u}$ and restricting the affinity to the latent-space only recover exactly the LC-TIM baseline.

\subsection{Optimization}
We minimize the objective function in Eq.~\eqref{eq:objective_ours} with an Alternating Direction Method (ADM) as in LC-TIM that yields to closed-form updates. The key is to introduce, for every query patch $i$, an auxiliary distribution $\mathbf{q}_i\in\Delta_K$ over classes. ADM then alternates two closed-form minimization until convergence (proof of convergence can be found in \cite{Li2026tim++}). \vspace{1mm}
\begin{itemize}
    \item \textbf{$q$-update}: It takes the same form as in the baseline LC-TIM except that $\bar p_i$ integrate spatial information and the denominator comes from the prior-anchored marginal entropy. Dividing the class marginal $\mu_k$ by $\hat\pi_k$ drives the predicted distribution towards the estimated prior (setting $\hat\pi_k$ uniform recovers the baseline LC-TIM).
    \begin{equation}
        q_{ik} \;\propto\;
        \frac{
            p_{ik}^{\,1+\alpha}\;\;
            \hat y_{ik}^{\,\beta}\;\;
            \bar p_{ik}^{\,\delta}
        }{
            \big(\mu_k / \hat\pi_k\big)^{1/2}
        },
        \qquad
        \mu_k = \sum_{i\in Q} p_{ik}^{\,1+\alpha}\,\hat y_{ik}^{\,\beta}\,\bar p_{ik}^{\,\delta},
        \label{eq:qupdate}
    \end{equation}
    followed by a normalization $q_{ik}\leftarrow q_{ik}/\sum_{k'}q_{ik'}$. \vspace{1mm}
    \item \textbf{W-update}: It is the same as the baseline LC-TIM. The marginal prior only enters the $q$-update. 
\end{itemize}

\section{Experiments}
\label{sec:exp}

\noindent \textbf{Datasets.} We use WSI datasets that provide a pixel-level segmentation mask, allowing us  to divide each WSIs into $448 \times 448$ patches with a patch-level label. 
BACH~\cite{Aresta2019bach} grades breast cancer into four classes. CATCH~\cite{Wilm2022catch} consists of canine skin tissue classified into 13 tissue types. SKINCANCER~\cite{Thomas2021skincancer} comprises 11 classes of skin tissue.  TIGER~\cite{2022tiger} is a binary dataset distinguishing breast cancer from healthy tissue. 
Note that CATCH and SKINCANCER are highly class-imbalanced, with several classes that can be absent from a WSI. \vspace{1mm}

\noindent \textbf{Annotations. } To define the set $\mathcal{S}$ of annotated patches, for each WSI, we draw a fixed number of shots randomly per class present in the slide, mirroring clinical practice where a pathologist is always involved in the diagnosis. \vspace{1mm}

\noindent \textbf{Implementation details. } To obtain the ZS predictions and the embeddings used in the CE, we use the VLM CONCH~\cite{Lu2024conch}; to obtain the embeddings used in the regularization and to estimate $\hat{\pi}$, we use the vision encoder UNI-2h~\cite{Chen2024uni}. 
We set $\lambda=0.4$, $\alpha=0.1$, $\beta=0.05$, $\delta=1$, $\tau=120$, $\kappa=4$, and $\sigma_s=448$. \vspace{1mm}

\noindent \textbf{Baselines. } We compare our method against the ZS predictions and the methods build upon: TIM++~\cite{Li2026tim++} and LC-TIM~\cite{Khoury2026lctim}, as well as $\alpha$-TIM~\cite{Veilleux2021alphatim}, a version of TIM~\cite{Dolz2020tim} designed for imbalanced scenarios. 
To assess whether the gain of our method stems from the knowledge of which classes are present in the slide, encoded by the UNI prior, we additionally report the ZS performance restricted to the set of classes present in the slide. 
To compare methods under class imbalance, we report macro-F1 and the F1 score for the minority class on the slide, in addition to accuracy, averaged over three random seeds, each seed corresponding to a different sampling of the support set $\mathcal{S}$.
\begin{figure*}[t!]
    \centering
    \includegraphics[width=\linewidth]{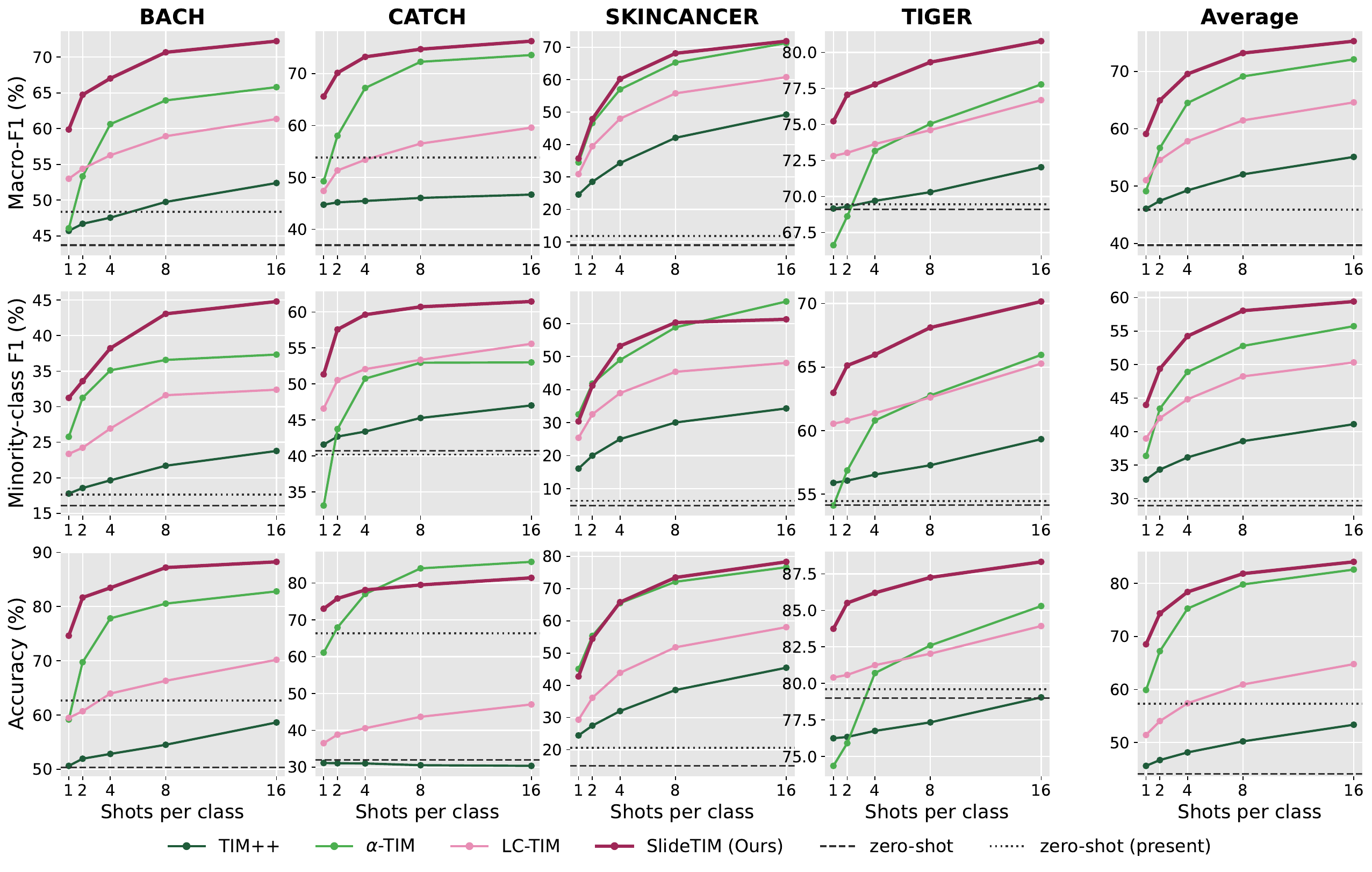}
    \caption{Macro-F1 score on each dataset, together with the F1 of the minority-class present in the slide and the accuracy over the four datasets. Results are averaged over three seeds. SlideTIM surpasses all baselines in macro-F1 and accuracy.} 
    \label{fig:main_results}
\end{figure*}

\subsection{Comparison with baselines}

The results are reported in Figure~\ref{fig:main_results}. 
Compared to ZS, the average macro-F1 improves by +19.4pp and +34.6pp, at one and 16 shot respectively, and the accuracy by +24.5pp and +40.0pp. 
When the classes absent from the slide are removed to compute the ZS (ZS present), the gain remains large, from +13.2pp at one shot to +29.4pp at 16 shots in macro-F1. This indicates that the improvement of our method does not merely stem from the knowledge of which classes are present in the slide, as encoded by the marginal prior. 
The minority-class F1 follows the same trend, confirming that our method predicts the rare class accurately without over-predicting it.
Among the baselines, $\alpha$-TIM performs best and TIM++ worst, yet our method still surpasses the best competitive baseline by +8.1pp in macro-F1 at one shot and $\alpha$-TIM by +3.2pp at 16 shots. 

A visual comparison of the best baseline and our method is shown in Figure~\ref{fig:visual_results}. 
While $\alpha$-TIM and SlideTIM handle the imbalanced setting, $\alpha$-TIM tends to over-predict the minority-class compared to SlideTIM that keeps the rare class confined to its actual regions while no longer predicting classes absent from the slide. 

\subsection{Ablation studies: Impact of additional terms}

\begin{table}[t]
    \centering
    \caption{Ablation of the added terms at 4 shots per class, averaged over the four datasets. Highlighted lines correspond to LC-TIM and \textbf{SlideTIM (Ours)}.}
    \label{tab:ablation}
    \resizebox{.48\textwidth}{!}{
        \begin{tabular}{cccccc}
            \toprule
            Latent & Spatial & Prior & Macro-F1 & Raw acc. & Balanced \\
            \midrule
             \rowcolor{pink} \cmark & \xmark & \xmark & 57.8 & 57.4 & 64.3 \\
            \xmark & \cmark & \xmark & 58.8 & 57.8 & 65.4 \\
            \cmark & \cmark & \xmark & 59.4 & 58.4 & 65.9 \\
            \xmark & \xmark & \cmark & 60.2 & 68.1 & 67.0 \\
            \rowcolor{burgundy} \cmark & \cmark & \cmark & \textbf{69.6} & \textbf{78.4} & \textbf{75.3} \\
            \bottomrule
        \end{tabular}
    }    
\end{table}

We evaluate the impact of the two terms added to the objective function: the spatial information and the marginal prior. 
The results are summarized in Table~\ref{tab:ablation}. 
On average, spatial consistency yields better results than latent-space consistency. Combining both into a single neighborhood performs best by benefiting from the two sources of information.  
Adding the marginal prior is beneficial as it steers the predicted proportions towards the classes present in the slide, leading to increased accuracy.
Beyond their individual contributions, the two added terms are complementary and reach their best performance when combined, our full method outperforming each single component variant by approximately +9.0pp in macro-F1.
\vspace{-2mm}

\section{Conclusion}
\label{sec:conclu}

We propose SlideTIM, an adaptation of LC-TIM to the setting of WSIs that present complex spatial organization and a high class imbalance. 
We demonstrate that the addition of a spatial consistency and a marginal prior helps to be robust to these conditions, showing an increase of +19.4pp in average F1-macro with only one shot, and +34.6pp with 16 shots, compared to ZS. 
However, we suppose that the set of annotated patches contains all classes present in the slide and that the pathologist has not missed a class. It should be interesting in future work to evaluate how our method acts in this scenario.
Nevertheless, our method shows that reliable patch-level segmentation can be achieved from the few annotations a pathologist naturally provides during diagnosis.

\clearpage

\section{Compliance with Ethical Standards}
This is a study for which no ethical approval was required.

\section{Acknowledgments}
\label{sec:acknowledgments}
 T. Godelaine and M. Dausort are funded by the MedReSyst project, supported by FEDER and the Walloon Region.
B. Gérin is funded by the Walloon region under grant No. 2010235 (ARIAC by DIGITALWALLONIA4.AI) and C. De Vleeschouwer by the Fonds de la Recherche Scientifique (FNRS).
Computational resources have been provided by the Walloon Region and the FNRS under convention 2.5020.11.

\bibliographystyle{IEEEbib}
\small\bibliography{refs}

\end{document}